\documentclass{article}

\usepackage{arxiv}

\usepackage[utf8]{inputenc} 
\usepackage[T1]{fontenc}    
\usepackage{hyperref}       
\usepackage{url}            
\usepackage{booktabs}       
\usepackage{amsfonts}       
\usepackage{amsmath}
\usepackage{amssymb}
\usepackage{nicefrac}       
\usepackage{microtype}      
\usepackage{lipsum}
\usepackage{algorithm}
\usepackage{algorithmic}

\usepackage{pifont}
\usepackage{graphicx}
\usepackage{float}
\usepackage{subfig}
\usepackage{changepage}

\usepackage{hyperref}
\usepackage{enumitem}

\hypersetup{
	colorlinks=true,
	linkcolor=blue,
	urlcolor=blue,
}
\DeclareMathOperator{\mean}{mean}

\begin{document}

\title{Depth-Aware Arbitrary Style Transfer Using Instance Normalization}  

\author{
	Victor Kitov\\
	Lomonosov Moscow State University, \\
	Plekhanov Russian University of Economics \\
	\texttt{v.v.kitov@yandex.ru}
	\thanks{Web page: \url{https://victorkitov.github.io/}}
	\And
	Konstantin Kozlovtsev	\\
	Lomonosov Moscow State University\\
	\texttt{ko-sova@yandex.ru} \\
	\And
	Margarita Mishustina \\
	Lomonosov Moscow State University\\
	\texttt{margarita\_mishustina\_112@mail.ru}
}

\maketitle          
\begin{abstract}
Style transfer is the process of rendering one image with some content in the style of another image, representing the style. Recent studies of Liu~~\textit{et al.}~(2017) show that traditional style transfer methods of Gatys~\textit{et al.}~(2016) and Johnson~\textit{et al.}~(2016) fail to reproduce the depth of the content image, which is critical for human perception. They suggest to preserve the depth map by additional regularizer in the optimized loss function, forcing preservation of the depth map. However these traditional methods are either computationally inefficient or require training a separate neural network for each style. AdaIN method of Huang~\textit{et al.}~(2017) allows efficient transferring of arbitrary style without training a separate model but is not able to reproduce the depth map of the content image. We propose an extension to this method, allowing depth map preservation by applying variable stylization strength. Qualitative analysis and results of user evaluation study indicate that the proposed method provides better stylizations, compared to the original AdaIN style transfer method.
	
\keywords{image processing \and image generation \and depth estimation \and instance normalization.}
\end{abstract}

\section{Introduction}
The problem of rendering an image (called the \textit{content image}) in a particular style is known as \textit{style transfer} and is a long studied problem in computer vision. Early approaches~\cite{gooch2001non,strothotte2002non,rosin2012image} used algorithms with human engineered features targeting to impose particular styles. 

In 2016 Gatys~\textit{et al.}~\cite{gatys2016image} proposed an algorithm of imposing arbitrary style taken from user defined \textit{style image} on arbitrary content image by using representations of images that could be obtained with deep convolutional networks. However their method needed a computationally expensive optimization in the space of images requiring several minutes of processing a single image of moderate resolution on powerful GPUs. 
Ulyanov \textit{et al.}~\cite{ulyanov2016instance} and Jonson \textit{et al.}~\cite{johnson2016perceptual} proposed a real-time style transfer algorithm by passing a content image through a pretrained fully convolutional transformer network. Their methods required training a separate transformation network for each new style. 
Work of Liu et. al (2017)~\cite{liu2017depth} highlighted an issue with traditional style transfer methods that they failed to reproduce the depth map of the content image, that was critical for human perception of the result. To address this issue they extended traditional methods~\cite{gatys2016image} and \cite{johnson2016perceptual} with a regularizer, forcing preservation of the depth map of the content image. This yielded significant improvement of style transfer rendering quality but required computationally complex algorithm, requiring either solving high dimensional optimization problem for each content-style pair or fitting a separate transformer network for each style.
Later architectures, such as AdaIN~\cite{huang2017arbitrary} and other (\cite{ghiasi2017exploring}, \cite{li2017universal}), allowed transferring arbitrary style without training a separate network but lacked rendering quality due to failure to preserve the depth map of the content image.

\begin{figure}
	\centering
	\includegraphics[width=1\textwidth]{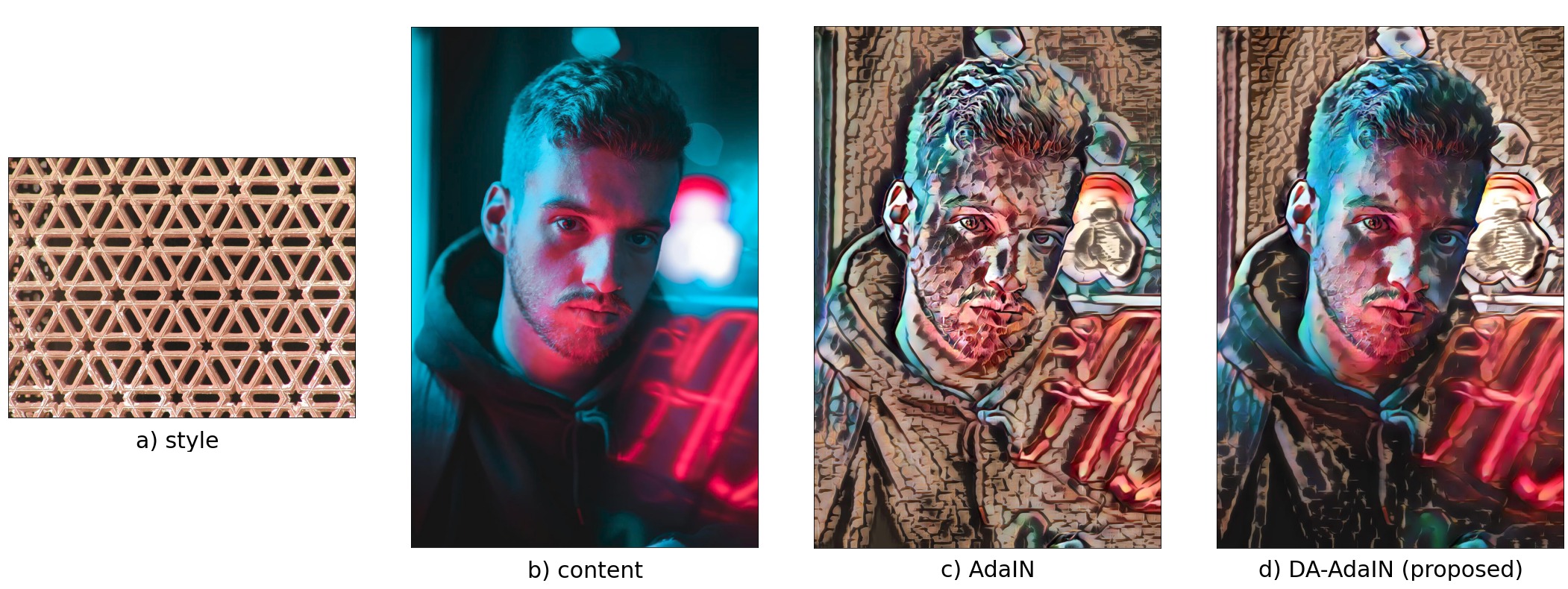}
	\caption{AdaIN and proposed Depth Aware AdaIN method comparison.}
	\label{fig:uniform_vs_depth_aware}
\end{figure}

In the work a depth aware AdaIN method extension (DA-AdaIN for short) is proposed that allows to preserve the depth map of the content during stylization, as shown on fig.~\ref{fig:uniform_vs_depth_aware}d by applying style with spatially variable strength: more close regions, standing for foreground, are stylized less, and more distant regions, standing for background, are stylized more. 

Qualitative analysis suggests that proposed modification leads to rendering quality improvement. User study confirms that proposed algorithm gives on average better results. 

The remainder of the paper is organized as follows. In section~\ref{sec:depth-estimation} two recent depth estimation methods are compared and the best one is used in later analysis. Section~\ref{sec:methods} describes standard AdaIN method and our proposed modification. Section~\ref{sec:evaluation} provides qualitative analysis of the proposed method, its dependency on major parameters and results of a user study, where AdaIN and proposed methods are compared. Finally section~\ref{sec:conclusion} concludes.

\section{Depth estimation}
\label{sec:depth-estimation}

\subsection{Methods}
Monocular depth estimation is a problem of finding a depth map $D\in\mathbb{R}^{W\times H}$ for arbitrary color image $I\in\mathbb{R}^{W\times H\times 3}$. Since $I$ is a color image, it has three channels, standing for red, green and blue color intensities. $D$ is a single channel image, having the same width $W$ and height $H$ as $I$, with $D(x,y)$ equal to the distance of pixel $I(x,y)$ to the camera. 

For the purposes of style transfer we are interested to discriminate between central objects, that are more close to the camera, from background objects, that are more distant. So absolute accuracy of depth prediction is not as important as relative accuracy.

We compare two recent methods for monocular depth estimation - \textit{monodepth2}~\cite{monodepth2} and \textit{MiDaS}~\cite{Ranftl2019}, using official public implementations of both. 

MiDaS is a supervised model in contrast to monodepth2, which is self-supervised, which means that it did not use true depth values during training. Monodepth2 was trained on single KITTI 2015 dataset \cite{menze2015object}, covering outdoor scenes, taken by the camera on the car. MiDaS was trained on five different datasets, covering indoor and outdoor scenes with static and dynamic objects in various contexts. 

MiDaS has a single realization, whereas monodepth2 has nine: mono-640x192, stereo-640x192, mono+stereo-640x192, mono-1024x320, stereo-1024x320, mono+stereo-1024x320, mono-no-pt-640x192, stereo-no-pt-640x192, mono+stereo-no-pt-640x192. They differ in training data used (mono, stereo or both), resolution of training data and weights initialization.

\subsection{Comparison}
\begin{figure}
	\centering
	\includegraphics[width=1.0\textwidth]{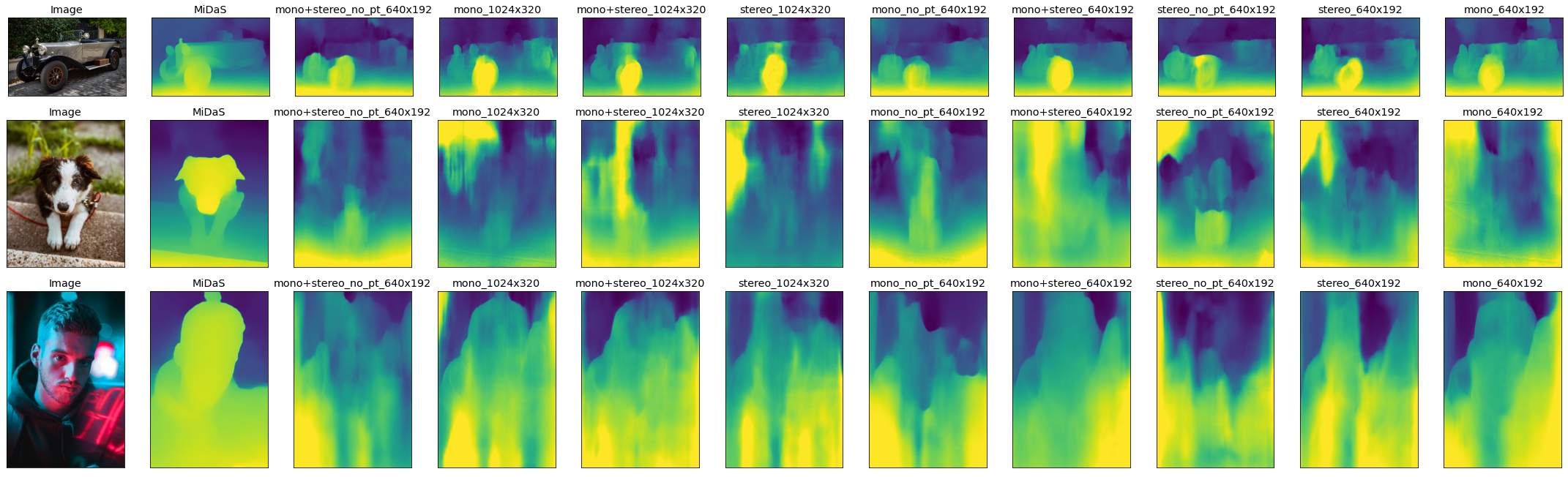}
	\caption{Qualitative comparison of \textit{MiDaS}~\cite{Ranftl2019} (2nd column) and different realizations of \textit{monodepth2}~\cite{monodepth2} depth estimation methods (columns 3-11). MiDaS is robust to different scenes, whereas monodepth2 has poor generalization for non-road objects.}
	\label{fig:depth_methods}
\end{figure}

Since style transfer may be applied to arbitrary images, we need a depth estimation method that is robust across different types of scenes. Qualitative check for random images shows significant superiority of the MiDaS method, as can be seen on fig.~\ref{fig:depth_methods}.

To compare methods quantitatively we apply them on the test subset of the DIW dataset\cite{chen2016single}, having diverse kinds of images. Neither of the methods used this dataset for training. The test subset used contains 73983 images with sparse labels: indicators for pairs of points, whether the first point is more distant or close to the camera, compared to the second point. Accuracy results are reported in table~ \ref{tab:depth_accuracy}. These results confirm that MiDaS is more accurate depth estimation model for images of general kind, so we will use this method in later analysis. This is an expected result since MiDaS was trained in supervised way on a number of diverse datasets.

\begin{table}[H]
	\centering
	\begin{tabular}{|l|c|}
		\hline		
		Method & Accuracy \\
		\hline
		\textbf{MiDaS} & \textbf{0.87} \\
		mono+stereo-1024x320 & 0.69 \\
		mono+stereo-640x192 & 0.70 \\
		mono+stereo-no-pt-640x192 & 0.65 \\
		mono-1024x320 & 0.70 \\
		mono-640x192 & 0.71 \\
		mono-no-pt-640x192 & 0.65 \\
		stereo-1024x320 & 0.67 \\
		stereo-640x192 & 0.66 \\
		stereo-no-pt-640x192 & 0.62 \\
		\hline
	\end{tabular} \medskip
	\caption{Relative depth prediction accuracy for MiDaS~\cite{Ranftl2019} and different realizations of monodepth2~\cite{monodepth2} depth prediction methods on the DIW test dataset~\cite{chen2016single}.}
	\label{tab:depth_accuracy}
\end{table}

\section{Style transfer}
\label{sec:methods}

\subsection{AdaIN Method}

AdaIN method~\cite{huang2017arbitrary} is a recent powerful style transfer method, allowing to stylize any content image $I_c$ by any style image $I_s$ in real-time without any complex optimizations. Stylization result $\hat{I}$ is obtained by
$$
\hat{I} = g(AdaIN(f(I_c), f(I_s))).
$$
where $f(\cdot)$ is an encoder (taken as first few layers of VGG-19~\cite{simonyan2014very}) and $g(\cdot)$ is a decoder, trained to match the encoder in producing good stylizations for representative set of content and style images. $\operatorname{AdaIN}(x, y)$ is a variant of instance normalization~\cite{ulyanov2016instance}, where instance normalization parameters are taken from the style image representation. 

Define encoder representations 
$$\begin{gathered}
x=f(I_c), \quad x \in \mathbb{R}^{C \times H_c \times W_c} \\
y=f(I_s), \quad y \in \mathbb{R}^{C \times H_s \times W_s} \\
\end{gathered}$$
Then $\operatorname{AdaIN}(x, y)\in\mathbb{R}^{C \times H_c \times W_c}$ and is defined as
\begin{gather}
    \operatorname{AdaIN}(x, y)_{cij} =
    \sigma_c(y) \left( \frac{x_{cij}-\mu_c(x)}{\sigma_c(x)}
    \right) + \mu_c(y),
    \label{eq:adain} \\
    \mu_c(x) = \frac{1}{H W}\sum_{i=1}^{H}\sum_{j=1}^{W}x_{cij}, 
    ~\sigma_c(x) = \sqrt{\frac{1}{H W}\sum_{i=1}^{H}\sum_{j=1}^{W}(x_{cij} - \mu_c(x))^2}\\
    i=1,2,...H_c;\quad j=1,2,...W_c;\quad c=1,2,...C; 
    \label{eq:moments}
\end{gather}

\subsection{Proposed Extension}
Standard AdanIN method applies style uniformly across content image. To improve rendering quality of style transfer we propose to apply style with different strength in different regions of content image depending on their proximity to the camera. Closer regions we consider foreground, that needs to be preserved more, so we stylize it less. And vice versa more distant regions we consider background, that can be stylized more.

Uniform stylization strength control can be controlled by hyperparameter $\alpha\in [0,1]$ in the following formula:
\begin{equation} \label{uniform_style_strength}
\hat{I} = g( \alpha f(I_c) + (1-\alpha) \operatorname{AdaIN}(f(I_c), f(I_s))),
\end{equation}
since $f(I_c)$ is the original unmodified content encoder representation, whereas  $\operatorname{AdaIN}(f(I_c), f(I_s))$ is fully stylized encoder representation.

Since we are interested in spatially variable strength control, we apply modified formula
\begin{equation}\label{eq:variable_strength_control}
\hat{I} = g( P \odot f(I_c) + (1-P)\odot \operatorname{AdaIN}(f(I_c), f(I_s))),
\end{equation}
where $P\in\mathbb{R}^{H_c \times W_c}$ is stylization strength map with strength values for each spatial position of content encoder representation and $\odot$ denotes element-wise multiplication repeated for every channel:
$$
\{P \odot F\}_{cij}=P_{ij}F_{cij}
$$
Algorithm~\ref{alg:weighting} shows steps for computing stylization strength map $P$ in formula~\ref{eq:variable_strength_control}. 
\begin{algorithm}
	\caption{Styliztation strength map estimation.}
	\label{alg:weighting}
\begin{algorithmic}[1]
	
\REQUIRE content image $I_c$, monocular depth estimation algorithm, size of content encoder representation $f(\cdot)$ $H_c\times W_c$, offset $\varepsilon\ge0$, prominence $\beta\ge0$.
\ENSURE Stylization strength map $P$.

\STATE Get depth map $D$ for content image $I_c$               \label{alg:depth}
\STATE Get proximity map $P=\max D - D$                        \label{alg:proximity}
\STATE Rescale $P$ to content encoder representation size $H_c\times W_c$
\STATE $P:=(P-\min P)/(\max P - \min P)$                       \label{alg:norm}
\STATE $P:=P-\mean P$                                          \label{alg:center}
\STATE $P:=1/(1+\exp(-\beta P))$                               \label{alg:sigmoid}
\STATE $P:=\min\{P,1-\varepsilon\}$                            \label{alg:constrain}
\end{algorithmic}
\end{algorithm}
MiDaS algorithm produces proximity map straight away, so for it steps~\ref{alg:depth},\ref{alg:proximity} are omitted. Max, min and mean operations are produced over all spatial positions and produce a scalar. Step~\ref{alg:norm} ensures that proximity has spread in [0,1] interval. Step~\ref{alg:sigmoid} controls contrast of the depth map by hyperparameter $\beta$:  higher $\beta$ corresponds to more prominent changes in the depth map around its mean and $\beta=0$ makes depth map constant converting proposed algorithm to standard AdaIN. Step~\ref{alg:constrain} constrains proximity map from above by $1-\varepsilon$. Hyperparameter $\varepsilon$ controls the minimal offset from the camera to regions on the image.

For stylization pre-trained AdaIN encoder/decoder~\cite{simonyan2014very,huang2017arbitrary} and  pre-trained depth network~\cite{Ranftl2019} is used. Computational advantage of our method is that it is learning-free: given pretrained encoder, decoder and depth estimation network, method does not require additional training for new styles. We name our algorithm \textit{Depth Aware Adaptive Instance Normalization} (\textit{DA-AdaIN} for short).

\section{Style Transfer Evaluation}
\label{sec:evaluation}

\subsection{Dependence on major parameters}
\begin{figure}
	\centering
	\includegraphics[width=0.45\textwidth]{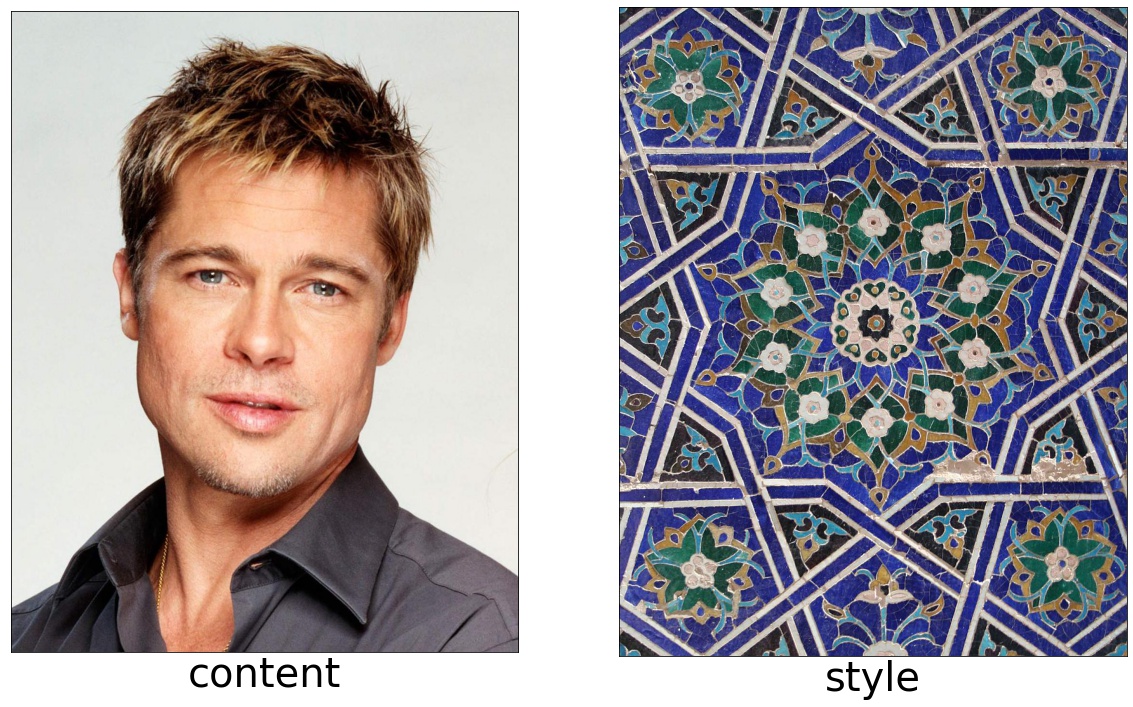}
	\caption{Style transfer result depending on depth contrast parameter $\beta$.}
	\label{fig:content_style}
\end{figure}

Proposed algorithm has two hyperparameters: $\beta>0$ controls prominence of proximity map around its mean value and $\varepsilon\in[0,1]$ controls minimal offset of the image regions from the camera. To study impact of these parameters on the stylization result we will use content and style images, shown on fig.~\ref{fig:content_style}

\begin{figure}
	\centering
	\includegraphics[width=1\textwidth]{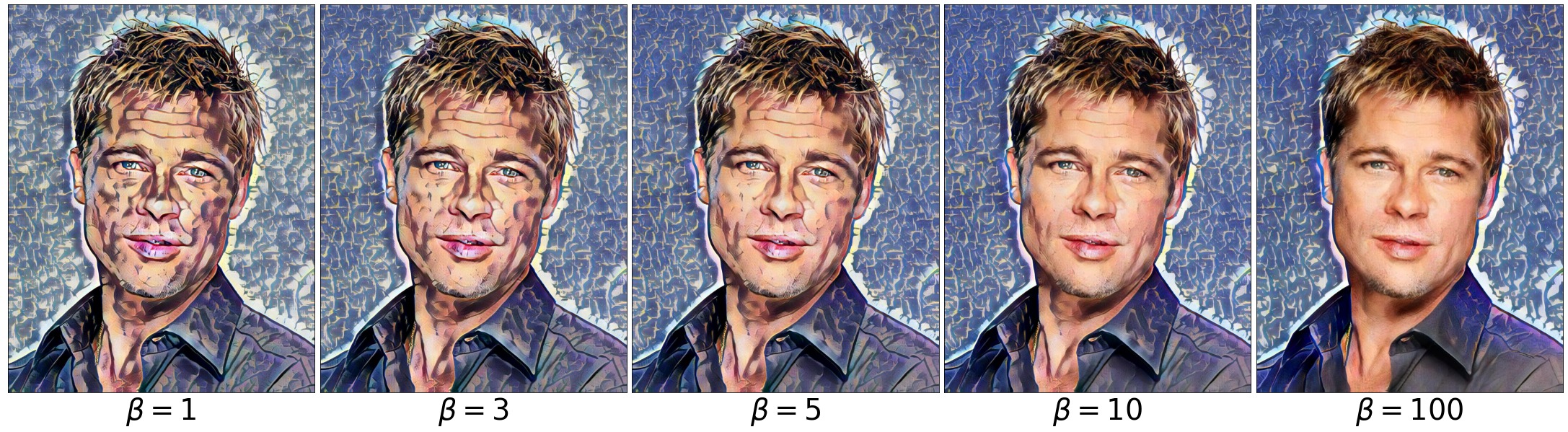}
	\caption{Style transfer result depending on depth contrast parameter $\beta$, $\varepsilon=0$.}
	\label{fig:beta_dependency}
\end{figure}

Fig.~\ref{fig:beta_dependency} shows how style transfer output depends on contrast parameter $\beta$. Higher values increase contrast (spread) of the proximity map values, while ensuring that they fall inside $[0,1]$ interval.

\begin{figure}
	\centering
	\includegraphics[width=1\textwidth]{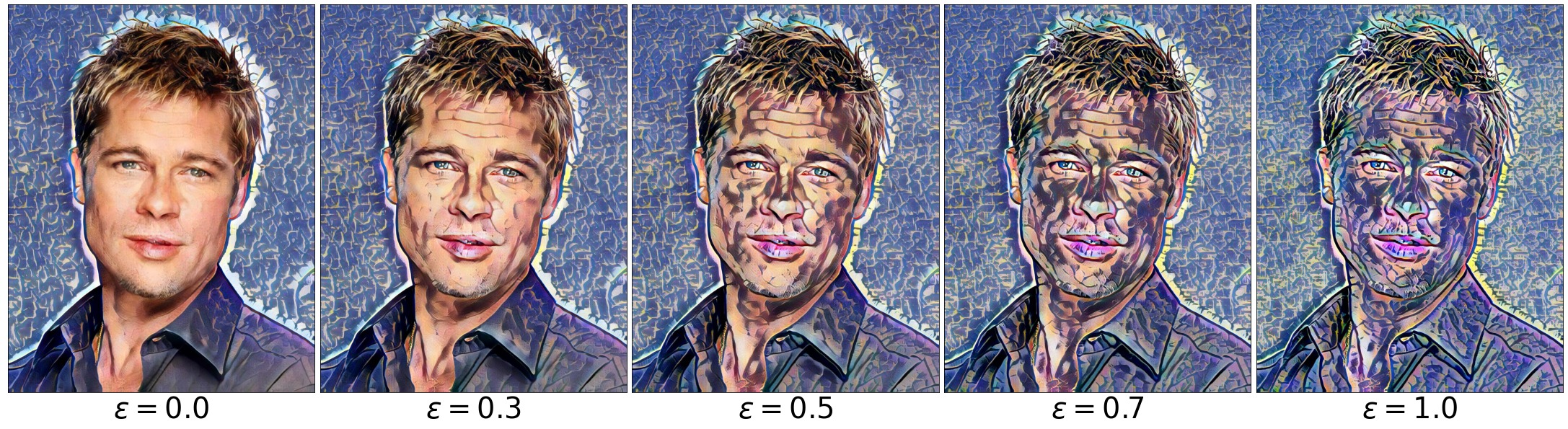}
	\caption{Style transfer result depending on proximity offset parameter $\varepsilon$, $\beta=20$.}
	\label{fig:eps_dependency}
\end{figure}

Fig.~\ref{fig:eps_dependency} shows dependency of style transfer results on proximity offset parameter $\varepsilon$. The lower is this offset, the closer proximity values may approach one in certain regions, forcing for that regions more content reconstruction and less style transfer. Higher values of $\varepsilon$ ensure that all image regions maintain certain distance from the camera and higher minimal impact of style transfer is ensured.

\subsection{Qualitative Comparison With AdaIN method}
\begin{figure}
	\centering
	\includegraphics[width=1\textwidth]{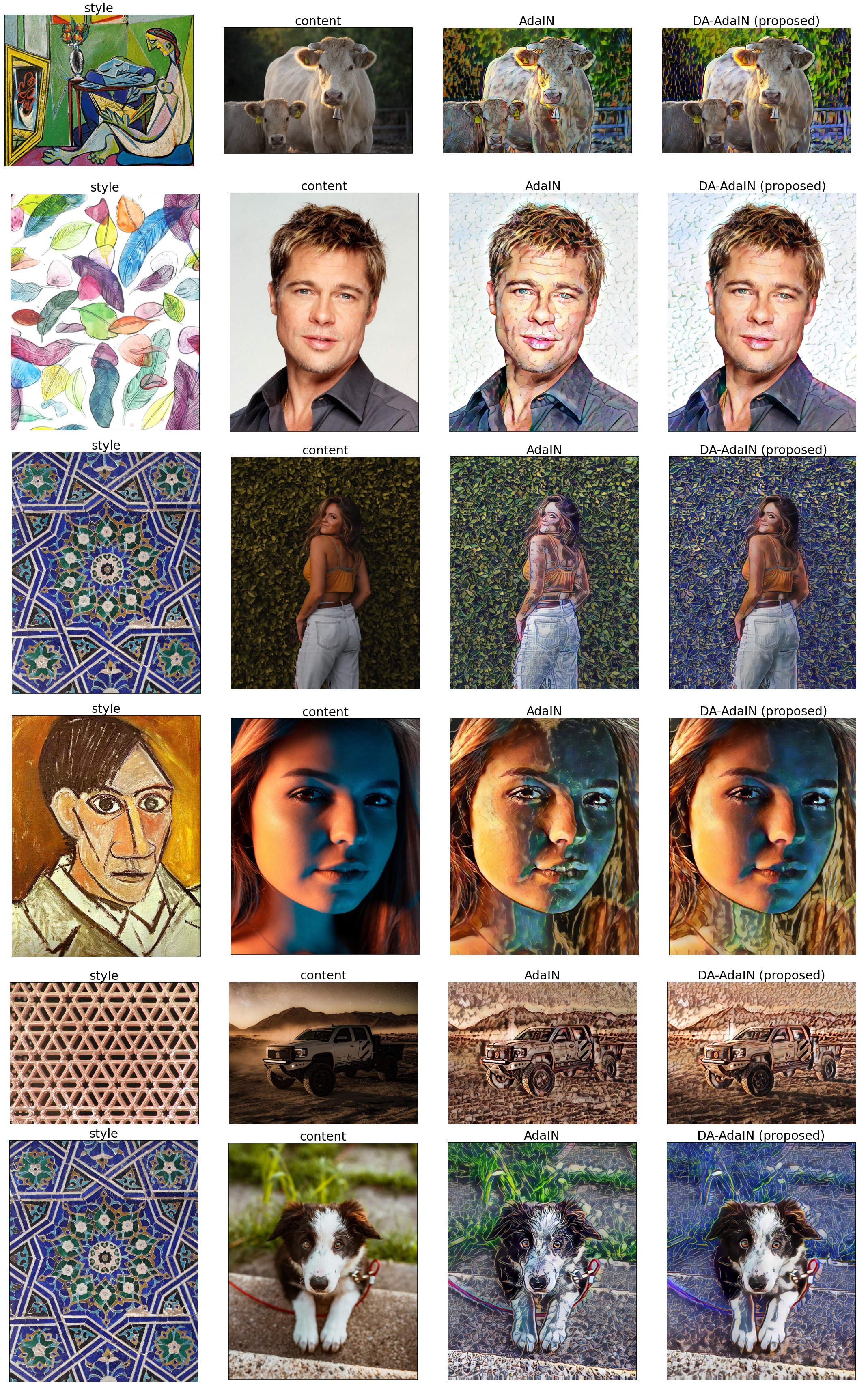}
	\caption{Comparison of style transfer results for AdaIN and proposed DA-AdaIN methods}.
	\label{fig:AdaIN_compare}
\end{figure}

Side-by-side comparisons of style transfer results by standard AdaIN method and proposed modification DA-AdaIN is visualized on fig.~\ref{fig:AdaIN_compare}. For DA-AdaIN we used $\varepsilon=0.15$ and $\beta=20$. Comparisons show that proposed method is capable to detect more close objects and highlight them by applying style transfer with smaller strength. More close objects generally are more important for the viewer and this strategy allows to preserve them better by less expressed stylization, which brings rendering improvement. However if this approach is used too strongly, it may create a noticeable disagreement between foreground and background rendering, as may be seen on the last row of fig.~\ref{fig:AdaIN_compare} where a huge proximity contrast between the foreground (the dog) and the background (the grass) forced the foreground look too photorealistic in strongly stylized context. To alleviate this issue we suggest to increase offset $\varepsilon$ or decrease contrast $\beta$.

\subsection{User Evaluation Study}

\subsubsection{Procedure.}
To provide a more general comparison of style transfer methods we conduct a user evaluation study, where 18 users were asked to pass a survey. The survey consisted of 20 image pairs, corresponding to stylizations by AdaIN and DA-AdaIN methods presented in random order, and the users had to select for each pair a stylization which they liked more. 360 responses were collected. For a set of different style and content images all possible stylizations were generated. Contents were selected to contain objects of different proximity to the camera, otherwise results between the two methods were indistinguishable.  A random subset of 20 results was selected for the survey. Contents were resized so that their smaller side is 1000 pixels and styles were resized to ensure that their smaller side is 300 pixels. We did not tell the respondents anything about the depth preservation concept and our algorithm details. For all inputs of DA-AdaIN we used $\varepsilon=0.15$ and $\beta=20$.

\subsubsection{Results.}
The results of user study evaluation study are presented on table~\ref{tab:compare}. Our method is preferred moderately more often than existing AdaIN method and this difference is statistically significant with 99\% confidence for exact binomial test.

\begin{table}[H]
    \centering
    \caption{Results of user evaluation study} \label{tab:compare}
    \begin{tabular}{r|c}
        Experiment & Ours vs AdaIN \\
        \hline
        \hline
        \# image pairs & 20 \\
        \# respondents & 18 \\
        \# responces & 360 \\
        \# votes for proposed method & 207  \\
        \textbf{proportion of votes for proposed method} & \textbf{57.5\%}  \\
        std. deviation of proportion & 2.6\%  \\
        p-value (exact binomial test) & 0.0026 \\
    \end{tabular} 
\end{table}

\subsubsection{Discussion.}
During the study it was found that DA-AdaIN was not very sensitive to proximity variability, so the contrast in variability had to be additionally increased by step~\ref{alg:sigmoid} of algorithm~\ref{alg:weighting}. For contents with very close objects to the camera, proximity became close to one for those objects and they received almost no stylization, so offset on step~\ref{alg:constrain} of algorithm~\ref{alg:weighting} was introduced to maintain certain guaranteed level of stylization for all parts of the image. These modifications ensured better rendering quality on average. $\beta=20$ and $\varepsilon=0.15$ are recommended. For particular content and style pair result may be improved even more by manual tuning of  $\beta$ and $\varepsilon$ parameters.

\section{Conclusion}
\label{sec:conclusion}
An extension to AdaIN method, allowing to preserve depth information from the content image, is proposed. All other benefits of AdaIN are preserved, namely fast real-time stylization and the ability to transfer arbitrary style at inference time without additionally training the model. Qualitative analysis reveals that the proposed method is capable to preserve information about proximity to the objects on the stylized image and results of the user evaluation study confirm that depth preservation is important for users, making them prefer our method more often than conventional AdaIN method.

\bibliographystyle{splncs04}
\bibliography{references}

\begin{thebibliography}{10}
\providecommand{\url}[1]{\texttt{#1}}
\providecommand{\urlprefix}{URL }
\providecommand{\doi}[1]{https://doi.org/#1}

\bibitem{chen2016single}
Chen, W., Fu, Z., Yang, D., Deng, J.: Single-image depth perception in the
  wild. In: Advances in neural information processing systems. pp. 730--738
  (2016)

\bibitem{gatys2016image}
Gatys, L.A., Ecker, A.S., Bethge, M.: Image style transfer using convolutional
  neural networks. In: Proceedings of the IEEE conference on computer vision
  and pattern recognition. pp. 2414--2423 (2016)

\bibitem{ghiasi2017exploring}
Ghiasi, G., Lee, H., Kudlur, M., Dumoulin, V., Shlens, J.: Exploring the
  structure of a real-time, arbitrary neural artistic stylization network.
  arXiv preprint arXiv:1705.06830  (2017)

\bibitem{monodepth2}
Godard, C., {Mac Aodha}, O., Firman, M., Brostow, G.J.: Digging into
  self-supervised monocular depth prediction  (October 2019)

\bibitem{gooch2001non}
Gooch, B., Gooch, A.: Non-photorealistic rendering. AK Peters/CRC Press (2001)

\bibitem{huang2017arbitrary}
Huang, X., Belongie, S.: Arbitrary style transfer in real-time with adaptive
  instance normalization. In: Proceedings of the IEEE International Conference
  on Computer Vision. pp. 1501--1510 (2017)

\bibitem{johnson2016perceptual}
Johnson, J., Alahi, A., Fei-Fei, L.: Perceptual losses for real-time style
  transfer and super-resolution. In: European conference on computer vision.
  pp. 694--711. Springer (2016)

\bibitem{li2017universal}
Li, Y., Fang, C., Yang, J., Wang, Z., Lu, X., Yang, M.H.: Universal style
  transfer via feature transforms. In: Advances in neural information
  processing systems. pp. 386--396 (2017)

\bibitem{liu2017depth}
Liu, X.C., Cheng, M.M., Lai, Y.K., Rosin, P.L.: Depth-aware neural style
  transfer. In: Proceedings of the Symposium on Non-Photorealistic Animation
  and Rendering. p.~4. ACM (2017)

\bibitem{menze2015object}
Menze, M., Geiger, A.: Object scene flow for autonomous vehicles. In:
  Proceedings of the IEEE conference on computer vision and pattern
  recognition. pp. 3061--3070 (2015)

\bibitem{Ranftl2019}
Ranftl, R., Lasinger, K., Hafner, D., Schindler, K., Koltun, V.: Towards robust
  monocular depth estimation: Mixing datasets for zero-shot cross-dataset
  transfer. arXiv:1907.01341  (2019)

\bibitem{rosin2012image}
Rosin, P., Collomosse, J.: Image and video-based artistic stylisation, vol.~42.
  Springer Science \& Business Media (2012)

\bibitem{simonyan2014very}
Simonyan, K., Zisserman, A.: Very deep convolutional networks for large-scale
  image recognition. arXiv preprint arXiv:1409.1556  (2014)

\bibitem{strothotte2002non}
Strothotte, T., Schlechtweg, S.: Non-photorealistic computer graphics:
  modeling, rendering, and animation. Morgan Kaufmann (2002)

\bibitem{ulyanov2016instance}
Ulyanov, D., Vedaldi, A., Lempitsky, V.: Instance normalization: The missing
  ingredient for fast stylization. arXiv preprint arXiv:1607.08022  (2016)

\end{thebibliography}

\end{document}